\documentclass[]{interact}

\usepackage{epstopdf}
\usepackage{subfigure}
\usepackage{amsmath}
\usepackage{amssymb}
\usepackage{subfigure}
\usepackage{algorithmicx}
\usepackage{bm}
\usepackage{float}
\usepackage{multirow}
\usepackage{stfloats}
\usepackage{algorithm}
\usepackage{algorithmicx}
\usepackage{algpseudocode}

\usepackage{natbib}
\bibpunct[, ]{(}{)}{;}{a}{}{,}
\renewcommand\bibfont{\fontsize{10}{12}\selectfont}

\theoremstyle{plain} 

\theoremstyle{definition}

\theoremstyle{remark}

\usepackage{color}

\begin{document}

\title{Multi-objective Optimization of Clustering-based Scheduling for Multi-workflow On Clouds Considering Fairness}

\author{
		\name{Feng Li \thanks{CONTACT Feng Li. Email: feng.li@ntu.edu.sg}, Wen Jun, Tan and Wentong, Cai}
		\affil{School of Computer Science and Engineering,
			Nanyang Technological University,
			Singapore 639798, Singapore}
	}
\maketitle

\begin{abstract}
Distributed computing, such as cloud computing, provides promising platforms to execute multiple workflows.
Workflow scheduling plays an important role in multi-workflow execution with multi-objective requirements.
Although there exist many multi-objective scheduling algorithms, they focus mainly on optimizing makespan and cost for a single workflow.
There is a limited research on multi-objective optimization for multi-workflow scheduling.
Considering multi-workflow scheduling, there is an additional key objective to maintain the fairness of workflows using the resources.
To address such issues, this paper first defines a new metric for fairness.
A multi-objective optimization model is designed based on makespan, cost, and fairness.
Then a global clustering-based multi-workflow scheduling strategy for resource allocation is proposed.
Experimental results show that the proposed approach performs better than the compared algorithms without significant compromise of the overall makespan and cost as well as individual fairness, which can guide the simulation workflow scheduling on clouds.
\end{abstract}


\section{Introduction}

Many applications in scientific and enterprise domains can be constructed as workflows which consist of tasks with dependencies~\citep{yu2007multi}.
To be able to process these computation or communication intensive tasks, workflows can be executed on the cloud.
Due to the changing market demand, different applications may need to be executed at the same time.
Hence, multiple simulation workflows may request cloud computing resources simultaneously, such that the execution of multiple workflows needs to be scheduled and coordinated across different applications. 

Currently, many works focus on the research of workflow scheduling algorithms, especially for single-workflow scheduling with or without quality-of-service (QoS) constraints, such as makespan or cost.
In existing literature, the makespan and cost are usually considered as the optimization objective.
They can be divided into \emph{heuristic-based} or \emph{metaheuristic-based} workflow scheduling ~\citep{masdari2016towards}.
However, it is still a great challenge to schedule multiple tasks from multi-workflow and simultaneously optimize multiple objectives for the following reasons:
(1) Scheduling objectives: Unlike single-workflow scheduling, multi-workflow scheduling should consider fair use of cloud computing resources by workflows in resource allocation. 
Hence, it is necessary to research the definition of fairness from the perspective of the multiple objectives, such as fairness in terms of execution time and cost.
(2) Scheduling ordering: For multi-workflow scheduling, it needs to consider not only task ordering within a single workflow but also task ordering among multiple workflows.
This reflect different priorities to determine resource instances for finishing task requirements in the order.
(3) resource allocation: clustering-based resource allocation has been researched in some papers which aim to reduce the communication time caused by data movement during scheduling.
But these cluster-based methods have not consider fairness in their optimization objectives.

In this paper, we focus on the scheduling of computing resource under multi-objective optimization for multiple workflows.
Particularly, we propose a multi-objective optimization model for multi-workflow considering the fairness of each workflow, and present cluster-based resource allocation approaches to improve optimization objectives and reduce the communication time.
The contributions of this paper are: (1) defining the fairness of multi-workflow scheduling considering the fairness of individual completion time and cost; (2) designing a cluster-based resource allocation strategy to reduce communication time; and (3) comparing and evaluating different scheduling algorithms in terms of multi-workflow multi-objective models.

The rest of this paper is organized as follows. Section~\ref{sec:review} summaries the related works. Section~\ref{sec:problem} presents the definition of scheduling problems, including multi-workflow models and multi-objective optimization models. Section~\ref{sec:algorithm} designs a clustering-based multi-workflow scheduling framework under multi-objective optimization, which aims to obtain balanced solutions among makespan, cost, and fairness. A comparative study of different clustering-based multi-workflow scheduling algorithms is presented in Section~\ref{sec:experiment}. Finally, we conclude the paper and provide insights on future directions.

\section{Literature Review} \label{sec:review}
Distributed deployment and execution of workflows rely on computing resources.
Hence, the scheduling of computing resource is an important consideration.
One of the main goals of workflow scheduling is to schedule the tasks of workflow(s) to resources according to the predefined objectives. Currently, many studies on workflow scheduling problem focus on a single-workflow scheduling. For example, \citet{lin2016pretreatment} proposed a cost-driven workflow scheduling strategy to minimize the execution cost of workflow while meeting the deadline requirement.
\citet{li2022wholistic} proposed a multi-factor scheduling strategy, including completion
time, energy consumption, and resource load balance for cloud-edge environment.
\citet{niu2020gmta} proposed a workflow allocation approach named geo-aware multiagent task allocation to optimize scientific workflow in clouds.
\citet{ahmad2016hybrid} proposed a hybrid genetic algorithm to optimize the makespan by using an efficient heuristic and modified the genetic operators.
In order to minimize the cost under deadline constraints, \citet{li2022mutation} proposed an improved farmland fertility algorithm combined with mutation and dynamic objective strategies.
\citet{li2022research} proposed a 3-stage scheduling optimization framework for executing IoT simulation workflows in a cloud.

With the popularity of modern high-performance computing platforms, workflow applications submitted by different users can run simultaneously on the same platform~\citep{tsai2015adaptive}.
Hence, scheduling algorithms need to be designed to process multiple workflows simultaneously. Currently, a handful of papers focus on multi-workflow scheduling and consider two main objectives, i.e., cost optimization under deadline constraints and execution time optimization under budget constraints~\citep{yu2006scheduling}. Most existing algorithms for multi-workflow scheduling under multiple criteria optimize only one objective while treating other criteria as constraints. Recently, some research investigated pure multi-objective approaches to optimize multiple criteria. Basically, makespan and cost are two primitive objectives in the workflow scheduling process. Multi-objective optimization algorithms are used to obtain balanced values between makespan and cost.
However, beyond makespan and cost, a specific issue of real concern for multi-workflow scheduling is the fairness of using resources for multiple workflows.
To generate a fair scheduling for given workflows, \citet{zhao2006scheduling} proposed two fairness policies to guide the selection of the workflow to be scheduled.
Fairness is defined based on the slowdown each workflow would experience when sharing resources with other workflows.
The next scheduled workflow will be determined based on the slowdown value. Although the fairness has been considered under single objective, it has not been studied for multi-objective optimization.

Workflow scheduling algorithms with multi-objective optimization can be solved using either heuristics or meta-heuristics approaches. Heuristics are only suitable for specific types of problems and different heuristics should be designed for different problems, whereas meta-heuristics can provide general solutions for different problems~\citep{chakravarthi2018workflow}. Meta-heuristics are more suitable to handle multi-objective optimization problems as they can easily obtain multi-objective trade-off solutions. Most scheduling algorithms based on meta-heuristic for multi-objective optimization were designed for the scheduling of a single workflow, makespan or cost being the key objective. \citet{zhu2015evolutionary} proposed an evolutionary algorithm to solve the scheduling problem of a single workflow to minimize makespan and cost. New schemes for problem-specific encoding and population initialization, fitness evaluation, and genetic operators were proposed in this algorithm.
\citet{hu2018multi} aimed to minimize makespan and cost while satisfying reliability constraints using particle swarm optimization.
\citet{chen2018multiobjective} proposed an ant colony algorithm based on a co-evolutionary population to optimize execution time and execution cost. 

Currently, clustering-based approaches have attracted much attention in the field of workflow scheduling over distributed computing resources. The initial purpose of task clustering was to reduce the data communication among tasks occurred in workflow execution. 
\citet{bittencourt2008performance} introduced a typical clustering-based heuristic scheduling approach named Path Clustering Heuristic (PCH). PCH adopts a clustering scheme to partition a workflow into several task groups. These tasks are then assigned to processors using a heuristic that minimizes the completion time. \citet{bochenina2015clustering} proposed a new algorithm named Multiple Deadline-constrained Workflows-Clustering-based method (MDW-C), implementing a clustering-based approach to schedule multiple workflows with soft deadlines.  

Although clustering-based methods are extensively used in heuristic-based scheduling to minimize transfer time, very few meta-heuristic-based multi-objective optimization approaches consider cluster-based scheduling. The above clustering-based schemes are designed for heuristic-based scheduling, where a heuristic-based resource allocation is performed after forming a cluster. However, this scheme is unsuitable for meta-heuristic-based multi-objective scheduling approaches because such methods determine the scheduling scheme from a global perspective and local heuristic information is not available in advance. Recently, some global clustering-based methods have been investigated to reduce the data movement of multi-objective optimization issues.
\citet{wangsom2019multi} proposed a peer-to-peer (P2P) clustering technique and compared it with the workflow scheduling without clustering. The key difference from the P2P approach is that it does not allow clustering between tasks with an undirected relationship.
To resolve the issue, Multilevel Dependent Node Clustering (MDNC) was proposed~\citep{wangsom2019multi1}, which can cluster several dependent nodes in different levels of workflows together.
The goal of the above two clustering-based multi-objective optimization approaches is to reduce data movement as well as not degrade parallelism by merging pipeline tasks into a single cluster. However, the effectiveness of clustering-based multi-objective optimization has not been verified in multi-workflow scheduling. 

\section{Problem formulation} \label{sec:problem}

\subsection{Multi-workflow model}
Computational tasks on the cloud are usually submitted as a workflow with precedence constraints. A workflow containing multiple computing tasks is executed on computing resources. Supposing $N$ workflows $W= \{W_1,...,W_N\} $ are to be scheduled, in which each workflow is represented by a direct acyclic graph (DAG) consisting of a set of tasks and dependencies between them. For instance, $W_g $ consists of a set of tasks $W_g^t = \{t_1,...,t_{n_g}\} $ and a set of directed edges $W_g^d = \{e_{ij}\mid t_i, t_j \in W_g^t \} $. An edge $e_{ij} $ represents a precedence relationship from task $t_i $ to task $t_j $, and the execution between them should follow the precedence relationship. All predecessors of task $t_j$ are denoted as:
\begin{eqnarray}
pred(t_j)=\{t_i\mid e_{ij} \in W_g^d\}
\end{eqnarray}
We define the entry set $t_{entry} $ of workflow $W_g $ satisfying:
\begin{eqnarray}
pred(t_{entry})=\emptyset
\end{eqnarray}

Different types of resources provided by the cloud platform are represented by a set of resources $R$. The purpose of multi-workflow scheduling is to map tasks from different workflows to resources in a way that meets objectives while preserving the precedence constraints among tasks within their workflows and coordinating the tasks between workflows.

\subsection{Multi-objective optimization model} 
Resource allocation for multi-workflow need to consider two different objectives simultaneously. One is to optimize global Quality of Service (QoS) metrics, such as minimizing overall completion time and execution cost. The other is to enable all workflows to use resources in a relatively fair way. Hence, fairness is also important for multi-workflow scheduling issues. These two objectives can have conflicting outcomes. The purpose of scheduling is to allocate tasks from different workflows onto cloud resources and balance these three objectives.

\subsubsection{Execution time and cost}

The overall complete time and execution cost of multi-workflow are based on the complete time and execution cost of each task, which are calculated by the following equations.

The execution time of task $t_i$ on resource $r$ in set $R$ is:
\begin{eqnarray}
	ET_{ir}=\frac{WL_{i}}{CU_{r}}  \label{aaa}
\end{eqnarray}

\noindent in which $WL_{i}$ denotes the workload of task $t_i$ and  $CU_{r}$ describe the CPU capacities of resource $r$.

The communication time between task $t_i$ and task $t_j$ is:
\begin{eqnarray}
	CT_{ij}^{rs}= \left\{
	\begin{array}{lll}
		0 & if & r=s \\
		\frac{DS_{ij}}{min(BW_r, BW_s)} & if & r\not=s
	\end{array}
	\right.  
\end{eqnarray}

\noindent$CT_{ij}^{rs}$ means the communication time for transferring data from task $t_i$ to $t_j$ executed on resource $r$ and $s$. $DS_{ij}$ is the data size transferred from task $t_i$ to $t_j$. $BW_r$ denotes the bandwidth of resource $r$. $CT_{ij}^{rs}$ is calculated by the ratio of $DS_{ij}$ to the minimum bandwidth between their execution resources. If two tasks are executed on the same resource, the communication time equals to zero.

The start time of task $t_i$ on resource $r$ is:
\begin{eqnarray}
	ST_{ir}= \left\{
	\begin{array}{lll}
		max(TR_{ir}) & if & t_i\in t_{entry}^g \\
		max \Big\{TR_{ir}, 
		\mathop{max} \limits_{t_j \in pred(t_i)} ( FT_{js}+CT_{ji}^{sr})  \Big\} & if & t_i\notin t_{entry}^g
	\end{array}
	\right.  
\end{eqnarray}
$ST_{ir}$ represents the start time of task $t_i$ on resource $r$. $TR_{ir}$ is the available time of resource $r$ for task $t_i$.  $ST_{ir}$ is determined by the available time of resource $r$, the maximum values of the sum of the finish time of its predecessors ($FT_{js}+CT_{ji}^{sr}$), and the communication time between its predecessors and itself.
Note that the start time is determined by the available time of resource if the task belongs to the entry task set.

The finish time of task $t_i$ on resource $r$ is:
\begin{eqnarray}
	FT_{ir}=ST_{ir} + ET_{ir}
\end{eqnarray}

\noindent$FT_{ir}$ is determined by the start time of the task $ST_{ir}$ and the execution time on a resource $ET_{ir}$.

The execution cost of task $t_i$ on resource $r$ is:
\begin{eqnarray}
	EC_{ir}=\frac{ET_{ir}\times CI_{r}}{BI_{r}}
\end{eqnarray}

\noindent$ CI_{r}$ represents the cost per interval unit when using resource type $r$, and $BI_{r}$ means the billing interval of resource type $r$.

Based on the execution time and cost per task, the overall complete time ($T_{total}$) and execution cost ($C_{total}$) are determined by:

\begin{eqnarray}
	T_{total}=\mathop{max}\limits_{{W_g\in W}}\Big\{ max_{t_i\in W_g^t, r\in R} (FT_{ir})\Big\}
	\label{t_total}
\end{eqnarray}
\begin{eqnarray}
	C_{total}=\sum_{{g}=1} ^ n \sum_{i=1} ^ {n_g} EC_{ir}
	\label{c_total}
\end{eqnarray}

\subsubsection{Fairness}

For the scheduling problem of multiple workflow, these workflows compete for the same resource set. That is, a workflow must share resources with other workflows.
In addition to optimize the overall makespan and cost, the fairness of each workflow should also be considered. As the time and cost are simultaneously considered in this paper, the fairness is related to makespan and cost of each workflow, which will be introduced below.

The definition of fairness is based on the loss rate. The loss rate is the sum of the slowdown and overspending (surcharge) that each workflow needs to experience, defined in \eqref{lossrate}.
Slowdown is the ratio of completion time when workflow $W_g$ is scheduled with other workflows and when it is scheduled alone ~\citep{zhao2006scheduling}, represented in \eqref{lossslow}.
Overcharging is the ratio of finish cost between these two schedule modes shown in \eqref{lossspending}.
The denominator in \eqref{lossslow} is the makespan when workflow $W_g$ keeps all available resources by itself. In the process of determining $FT_{is}(S)$, the objective is to minimize the finish time using the Heterogeneous Earliest Finish Time (HEFT) strategy.
The numerator in \eqref{lossslow} is the makespan of the same workflow scheduled with all other workflows in the set $W$, and the scheduling objective is to balance makespan and cost. The definition of overspending for workflow $W_g$ shown in \eqref{lossspending} is similar. The denominator is the cost of executing the same workflow on the cheapest resources whereas the numerator is the cost when scheduling the same workflow with all other workflows in the set $W$ under the objective of balancing makespan and cost. 

\begin{eqnarray}
	loss_{w_g}=loss_{w_g}(slowdown) + loss_{w_g}(overspending)
	\label{lossrate}
\end{eqnarray}

\begin{eqnarray}
	loss_{w_g}(slowdown) =\frac{max_{t_i\in W_g^t, r\in R} (FT_{ir}(M))}{max_{t_i\in W_g^t, s\in R} (FT_{is}(S))} 
	\label{lossslow}
\end{eqnarray}

\begin{eqnarray}
	loss_{w_g}(overspending)= \frac{\sum_{t_i\in W_g^t, r\in R}  (EC_{ir}(M))}{\sum_{t_i\in W_g^t, s\in R} (EC_{is}\!{(S)}) }
	\label{lossspending}
\end{eqnarray}

Fairness indicates that the loss rate of each workflow in the workflow set is as similar as possible. Consequently, unfairness implies that different workflows have significant variability in their respective slowdowns and overspendings. We use the standard deviation of the loss rates of all workflows as a metric of fairness. The unfairness is calculated as follows:
\begin{eqnarray}
	UF=\sqrt {\frac{{\sum_{g=1} ^{N}(loss_{w_g}-\overline{loss})^2}}{N}}
\end{eqnarray}
in which $\overline{loss}$ is the average loss rate value of all workflows in $W_g$. A low unfairness value indicates a small variance in loss rates between workflows. That is, scheduling is relatively fair to each workflow in terms of the considered objectives.

Finally, the purpose of scheduling is to find the mapping between tasks in workflows and resources, and to determine the occupied timing of each task for the mapping resource with the objectives shown in \eqref{eq}, in which the total time, cost, and unfairness should be minimized.
\begin{eqnarray}
	min \quad f=min (T_{total}, C_{total}, UF)
	\label{eq}
\end{eqnarray}

\section{Clustering-based multi-workflow scheduling algorithms} \label{sec:algorithm}

Clustering-based scheduling strategies have been applied to minimize communication time.
To schedule multiple workflows simultaneously, our goal is to cluster tasks that cannot be executed concurrently.
PCH is one of such methods and is extensively used in heuristic-based scheduling approaches. It uses a Depth-First Search (DFS) approach on a workflow to select the task $t_i$ with the highest priority $P_i$ and the earliest start time $EST_i$, and add it to a cluster, until it reaches a task with an unscheduled predecessor. Once a cluster forms, PCH selects a resource that can execute the current cluster in the fastest time.
For PCH, local task execution information is used to choose tasks to be clustered.
In other words, to obtain the value of $EST_i$, resource allocation must be performed immediately after the creation of a cluster to determine where the current cluster will be placed.
This strategy is unsuitable for global multi-objective optimization as local information is not available in advance.
Hence, we attempt to adapt the current DFS-based clustering and propose a new clustering-based scheduling framework. In the framework, approaches for global information-based task clustering and resource allocation are proposed. That is, resources are allocated after all tasks from different workflows are clustered, which is different from the PCH approach.
The proposed scheduling strategies for multiple workflows are uniquely determined by: (1) the method of task clustering; (2) the method of task or cluster ordering from different workflows; (3) the method of clustering-based resource allocation.
Each step is described in detail below: 

\subsection{Task clustering and ordering}
The major difference between cluster-based heuristic scheduling for single-objective optimization and cluster-based meta-heuristic scheduling for multi-objective optimization is that resource allocation for multi-objective optimization based on the meta-heuristics starts only after all tasks have been clustered and ranked, unlike heuristic-based resource allocation for single-objective optimization which is performed at the time of each cluster formation.

To implement cluster-based resource allocation for multi-objective optimization of multi-workflow, we present a DFS-based approach to cluster tasks in each workflow, which is named Depth-First-Search considering Critical Subsequent Task (DFS-CST). For DFS-CST, the first task selected to form a cluster $C_k$ is the unscheduled task with the highest priority, and the next problem is that how to select the subsequent tasks to be clustered based on DFS. In \eqref{eqcst}, task $t_i$ and its successor $t_k$ are clustered together where $t_j$ is a successor of $t_i$. $t_k$ represents the critical successive task of task $t_i$, which is the unclustered successor task of $t_i$ with the largest sum of its average computation time $\overline{ET}_j$ plus its average communication time $\overline{CT}_{ij}$. This method does not rely on local information and thus can be used for global-based clustering strategies. Before resource selection, all tasks should be clustered based on DFS-CST.
\begin{eqnarray}
	t_k =arg(\max _{t_j \in succ(t_i)} (\overline{CT}_{ij} + \overline{ET}_j))	
	\label{eqcst}
\end{eqnarray}

After clustering, one of the most followed important steps in multi-workflow scheduling is to decide how to sequence different tasks or clusters belonging to different workflows. In this paper, we use an interleave-based task ordering method, which indicates clusters belonging to different workflows in the same workflow set are selected in turns and only one task in the cluster is selected based on topology in each round. 

\subsection{Resource selection and task execution}
After the assignment of all tasks to the cluster is complete, the resource selection for task execution is followed. Tasks in the same cluster will be scheduled to the same resource, which is selected based on scheduling algorithms. For the process of resource selection, it determines the resource to execute tasks in each cluster with the objective of achieving a tradeoff among makespan, cost, and fairness. In order to compare different clustering methods fairly, we use the same multi-objective optimization algorithm framework, i.e., NSGA-III proposed by~\citet{deb2013evolutionary}, to optimize these three objectives. 

\begin{figure*}[t]
	\centerline{\includegraphics[width=0.8\textwidth]{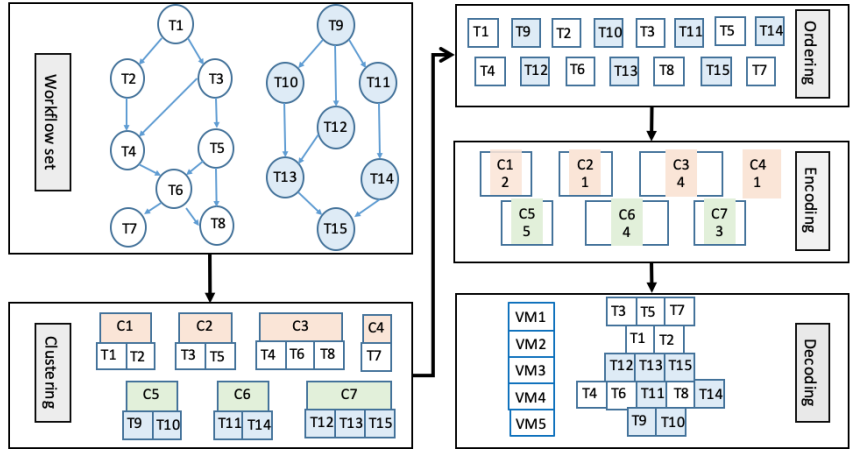}}
	\caption{An example of multi-workflow scheduling framework.}
	\label{fig:example}
\end{figure*}

Figure~\ref{fig:example} presents a case of multi-workflow scheduling. Taking a workflow set consisting of two workflows as an example, the clustering-based scheduling process is shown in this figure. The tasks in the workflow set are first clustered into seven clusters using the clustering strategy mentioned above. The clustering sequence does not represent the executing sequence on resources. In the step of ordering, we use interleave-based task ordering to rank tasks to be executed. They sort tasks from different workflows based on the interleave method. Then, the encoding process for resource allocation follows. The integer values in the encoding step indicate the types of resources selected for the corresponding clusters. For instance, the first value of the encoding step in Figure~\ref{fig:example} is 2, which means that tasks $t_1$ and $t_2$ in cluster 1 are allocated to resource 2 for execution. At last, a decoding process is required to obtain a scheduling plan. It combines the ordering step and the encoding step to determine where the tasks are to be executed and when they are to be scheduled. During the decoding process, the tasks shown in each resource do not represent the execution period, but rather denote their scheduling sequence on the chosen resource. The real start time and finish time of a task are determined not only by the execution sequence on the chosen resource, but also by the finish time of its predecessors.

\begin{algorithm}[ht!]
	\caption{Multi-workflow scheduling algorithms based on NSGA-III.}
	\label{alg:DFS-NSGA-III}
	\begin{algorithmic}[1]
		\Require
		A set of workflows,  $W= \{W_1,...,W_N\}$; A set of tasks $W_g^t = \{t_1^g,...t_{n_g}^g\} $ and a set of directed edges $W_g^d = \{e_{ij}^g\mid t_i^g, t_j^g \in W_g^t \} $ for each workflow $W_g$; Total number of tasks is $T_n = \sum_{g=1}^{n}{n_g}$; A set of resources, $R= \{r_1,...,r_m\} $; Number of trade-off solutions, $K$.
		\Ensure
		Set of $K$ trade-off schedules based on objectives. $S = \{S_k | k = 1, \cdots K\}$, where each $S_k$ is a schedule (i.e., a set of ($t_i$, $r_j$) tuples).
		\Statex\ //Task clustering
		\State k =0
		\For{each $W_g\in W$}
		\While{not each task in $ W_g$ has been clustered} 
		\State $C_k$ $\Leftarrow$ $\emptyset$
		\State $t$ $\Leftarrow$ unscheduled node with highest priority;  
		\State $C_k$ $\Leftarrow$ $C_k \cup t$;
		\While{$t$ has unscheduled successors}
		\State $t_{succ} \Leftarrow$ selected successor of $t$ using  
		\State \qquad DFS-CST;
		\State $C_k \Leftarrow C_k \cup t_{succ} $ ;
		\State $t \Leftarrow t_{succ}$;
		\EndWhile
		\State k = k + 1;
		\EndWhile
		\EndFor\
		\Statex\ // Task-based ordering	
		\State Using task interleave-based ordering  
		\Statex\ // Clustering-based resource allocation
		\State Generate the initial solutions $S$ with $N$ randomly 
		\State \qquad generated populations for clusters as parents;
		\While{termination criteria not fulfilled}
		\State $S^{'} \Leftarrow$ Tournament selection operation for $S$;
		\State $S^{'} \Leftarrow$ Crossover operation for $S^{'}$; 
		\State $S^{'} \Leftarrow$ Mutation operation for $S^{'}$;
		\State $S_{t}\Leftarrow$ Combination of solutions $S$ and $S^{'}$;
		\State Decoding the resource allocation for each cluster to 
		\State \qquad each task;
		\State Evaluating objective functions (14) and ranking the 
		\State \qquad schedule set $S_{t}$ by nondominated sorting;
		\State Niche-preserving operation: Choose $K$ solutions from 
		\State \qquad the sorting results and update each $S_k$ in $S$;
		\EndWhile\\		
		\Return $S$
	\end{algorithmic}
\end{algorithm}

The detailed algorithm steps of cluster-based multi-workflow scheduling algorithms based on NSGA-III are shown in Algorithm~\ref{alg:DFS-NSGA-III}. Tasks in each workflow are clustered by different clustering techniques as described in the above subsection. For each workflow in the workflow set, if some tasks in the workflow are not clustered, a new cluster $C_k$ is built (lines 1-4). The first task selected to compose a cluster $C_k$ is the unscheduled task with the highest priority (lines 5-6). Then, its suitable successor is added to the same cluster based on the clustering strategy DFS-CST (lines 7-10), and the current task is substituted by the successor (line 11). Consequently, the length of a chromosome is equal to the number of clusters. If a task cannot be clustered into any existing cluster, it can be treated as an independent cluster. The ordering step should be conducted before resource allocation. Tasks from different workflows are ordered based on the interleave method (line 16). Once the clustering and ordering is complete, each cluster is allocated to a resource determined by NSGA-III. First, initial solutions are randomly generated (line 17-18), and then the operators of selection, crossover, and mutation are used (lines 20-23). Clusters in the chromosome are expanded to the full length of tasks based on decoding, and the targets are evaluated (line 24-27), and then, the best $K$ solutions are selected (line 28-29). These steps are iterated multiple times, and the final solutions $S$ are formed.

\section{Experiments} \label{sec:experiment}
In this section, experiments via numerical simulations are presented to evaluate different clustering approaches and to validate the tradeoff among makespan, cost, and fairness of multi-workflow scheduling. The proposed DFS-based global clustering approach is compared to two other clustering approaches, i.e., P2P-based clustering and MDNC-based clustering, which can also be used for global multi-objective optimal scheduling.

\subsection{Performance Evaluation Metrics}
Currently, two performance metrics are extensively adopted to evaluate multi-objective optimization, i.e., Inverted generational distance (IGD)~\citep{liu2013decomposition} and hypervolume (HV)~\citep{jiang2017strength}. Although both indicators can reflect the convergence and diversity of solutions, the references they adopt are different. They have their own advantages and disadvantages. Hence, these two measures complement each other and can be used to evaluate the tradeoff fronts generated by various approaches, and they are described in detail below.

\subsubsection{Inverted generational distance}

IGD evaluates the proximity between the optimal solutions obtained by the proposed methods and the true Pareto front. The true Pareto optimal solutions of the simulation problems are hard to obtain. In order to find the approximate optimal solutions, we merge all solutions obtained by all proposed algorithms into a union set. Then, the solutions that are not dominated by any other solutions in the union set are taken as the final optimal solutions. The formulation of IGD is given in \eqref{eq5}:

\begin{eqnarray}
	IGD(P, P^{*})=\frac{(\sum_{x\in P^{*}} min_{y \in P} dis(x,y))}{\left| P^{*} \right|}
	\label{eq5}
\end{eqnarray}
where $P$ is the solution obtained by one of the five methods and $P^{*}$ is the true Pareto optimal solution. $dis(x,y)$ denotes the Euclidean distance from point $x$ in $P^{*}$ to the solution $y$ in $P$. The smaller the IGD value, the better solutions are found. 

\subsubsection{Hypervolume}

HV indicates the volume of the objective space covered between the obtained Pareto front and a reference point~\citep{zitzler2003performance}, which is given in \eqref{eq6}. $v_i$ is the hypervolume built by the reference point and solution $i$, and $X$ is the number of solutions. A larger HV value is preferred, indicating the solution set is well distributed. 
\begin{eqnarray}
	HV=\delta (\cup_{i=1}^{\left| X \right|} v_i)
	\label{eq6}
\end{eqnarray}

\subsection{Experiment based on Benchmark Workflows}
In order to objectively evaluate different clustering strategies, five benchmark workflows generated by the simulator described in \citet{chen2012workflowsim} are first chosen as the basic dataset.
They are \emph{Montage}, \emph{Epigenomic}, \emph{Inspiral}, \emph{CyberShake}, and \emph{Sipht}.
To simulate multi-workflow scenarios, two, three, four, and five workflows are randomly selected from these five benchmark workflows to form four different dataset, respectively.  

Based on the above workflow sets, the proposed DFS-CST is evaluated against other methods, P2P and MDNC, to solve the proposed multi-workflow multi-objective scheduling. They are integrated into the NSGA-III framework. The parametric values of NSGA-III are set as: population size $n = 50$, generations $g = 200$, crossover rate $cr = 0.8$, and mutation rate $mr = 0.01$.
The length of a chromosome is determined by the number of clusters. Each experiment is repeated 10 times due to the randomness of the algorithm.
The performance of different algorithms is analyzed based on IGD and HV. 

\begin{figure*}
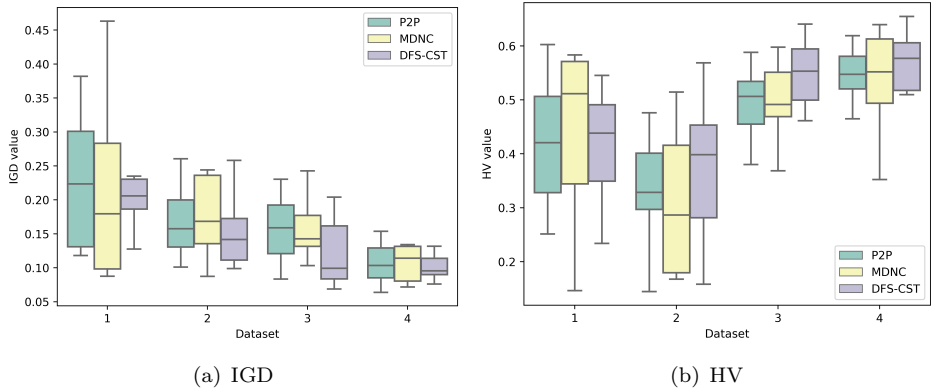

	\centering
	\subfigure[IGD]{\includegraphics[width=0.43\textwidth]{FIG_list/igd_for_dataset_three_method_real_world.pdf}}
	\subfigure[HV]{\includegraphics[width=0.43\textwidth]{FIG_list/hv_for_dataset_three_method_real_world.pdf}}
	\caption{Box plots of IGD and HV of each algorithm on benchmark workflows.}
	\label{real_world-igd-hv}
\end{figure*}

\begin{figure}[t]
	\centerline{\includegraphics[width=0.43\textwidth]{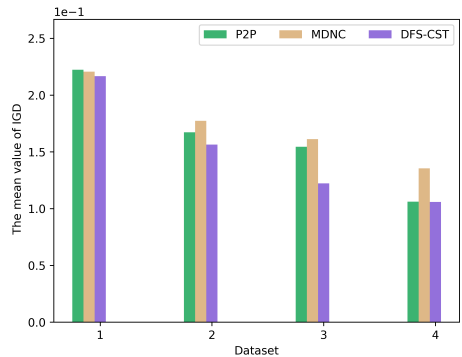}}
	\caption{Mean IGD of each algorithm on benchmark workflows.}
	\label{igd_real_world}
\end{figure}

Fig.~\ref{real_world-igd-hv} shows the box plots of IGD and HV for the four datasets consisting of benchmark workflows.
The results of the IGD average values in ten runs are shown in Fig.~\ref{igd_real_world} and the average values and standard deviations of HV are shown in Table~\ref{hv_real_world}.
Smaller IGD means that the corresponding algorithm is better.
From these figures, it can be seen that DFS-CST performs best compared to the compared methods.   
Meanwhile, the larger the better for HV indicator.
From the figure and the table, it can be seen that DFS-CST can obtain the largest HV in three out of four cases with smaller standard deviations. 
Based on IGD and HV results, it can be concluded that DFS-CST outperforms the other two compared methods for benchmark workflows.

\begin{table*}
	\footnotesize
	\centering
	\caption{The average values and standard deviations of HV obtained by the three algorithms in the benchmark workflow set}
	\begin{tabular}{ccc|cc|cc}
		\hline
		\hline
		 \textbf{Real}&\multicolumn{2}{c}{P2P} & \multicolumn{2}{c}{MDNC} & \multicolumn{2}{c}{DFS-CST}\\
		\cline{2-7} 
		 \textbf{datasets} & \textbf{\textit{AVG.}}& \textbf{\textit{STD.}}& \textbf{\textit{AVG.}}& \textbf{\textit{STD.}}&\textbf{\textit{AVG.}}& \textbf{\textit{STD.}}\\
		
		\hline
1	&0.4282	&0.1169	&\textbf{0.4344}	&0.1635	&0.4179	&\textbf{0.0939}	\\
2	&0.3371	&\textbf{0.0875}	&0.3075	&0.1287	&\textbf{0.3692}	&0.1326	\\
3	&0.4955	&0.0618	&0.4851	&0.0843	&\textbf{0.5535}	&\textbf{0.0593}	\\
4	&0.5487	&\textbf{0.0442}	&0.5215	&0.1214	&\textbf{0.5574}	&0.0786	\\
		\hline
	\end{tabular}
\label{hv_real_world}
\end{table*}

\subsection{Experiment based on Synthetic Workflows}
\subsubsection{Setup}
Different workflows denote different simulation processes for different applications. In order to widely simulate different kinds of workflows, we use synthetic workflows to validate the performance of different clustering algorithms. The workflows are generated by four factors, i.e., the length of workflow sets (the number of workflows in a set), the number of tasks in each workflow, the communication to computation ratio of tasks, and the degree of parallelism of tasks in a workflow. Each factor has two levels, which are explicitly introduced below.

Firstly, two workflow sets with 5 and 30 workflows in each set are considered. Secondly, we assume that the number of tasks in each workflow is randomly generated based on two uniform distributions in the range of 10-20 and 40-60, respectively. Two ratios of communication data size to the ranges of computation workloads (CCR) in a workflow, i.e., 0.1 and 1000, are considered. 0.1 and 1000 represent the target workflow is computationally intensive and non-intensive, respectively. At last, two levels of parallelism, 5\% and 30\%, are considered. It denotes the range of the number of tasks in each layer in a workflow. For example, 5\% represents the range of the number of tasks that can be executed in parallel in each layer is 5\% of the total number of tasks in each workflow. 5\% and 30\% represent low and high degrees of parallelism, respectively. The four factors, each with two levels, can form sixteen workflows as shown in Table~\ref{tab:data}.
\begin{table*}
	\footnotesize
	\centering
	\caption{Datasets design}
	\begin{tabular}{|c|c|c|c|c|}
		\hline
		\multirow{3}{*}{Datasets}  & \multicolumn{4}{c|}{Workflow structures}\\
		\cline{2-5}
		& Number of
		tasks & Number of workflows &CCR & Parallelism degree \\
		\hline
		1& 10-20 & 5	&0.1	&0.05 \\
		\hline
		2 &10-20 & 5	&0.1	&0.3	\\
		\hline
		3 &10-20 & 5	&1000	&0.05\\
		\hline
		4 &10-20 & 5	&1000	&0.3\\
		\hline
		5& 10-20 & 30	&0.1	&0.05 \\
		\hline
		6 &10-20 & 30	&0.1	&0.3	\\
		\hline
		7 &10-20 & 30	&1000	&0.05\\
		\hline
		8 &10-20 & 30	&1000	&0.3\\
		\hline
		9& 40-60 & 5	&0.1	&0.05 \\
		\hline
		10 &40-60 & 5	&0.1	&0.3	\\
		\hline
		11 &40-60& 5	&1000	&0.05\\
		\hline
		12 &40-60 & 5	&1000	&0.3\\
		\hline
		13& 40-60 & 30	&0.1	&0.05 \\
		\hline
		14&40-60 & 30	&0.1	&0.3	\\
		\hline
		15 &40-60 & 30	&1000	&0.05\\
		\hline
		16 &40-60 & 30	&1000	&0.3\\
		\hline
	\end{tabular}
	\label{tab:data}
\end{table*}

\begin{figure*}
	\centering
	\subfigure[datasets 1-4]{\includegraphics[width=0.43\textwidth]{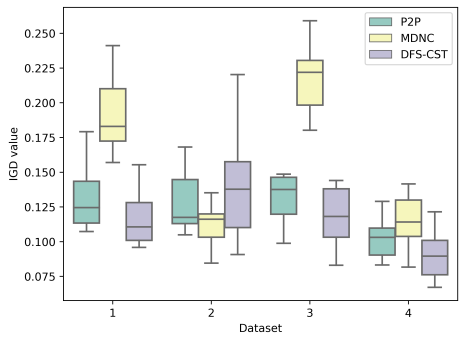}}
	\subfigure[datasets 5-8]{\includegraphics[width=0.43\textwidth]{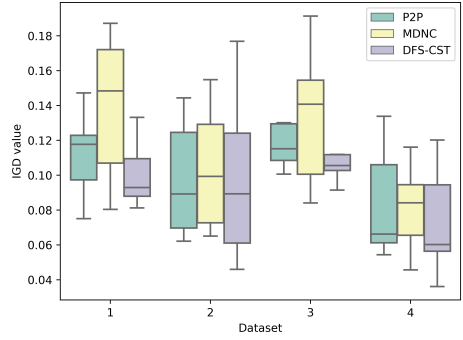}}
	\subfigure[datasets 9-12]{\includegraphics[width=0.43\textwidth]{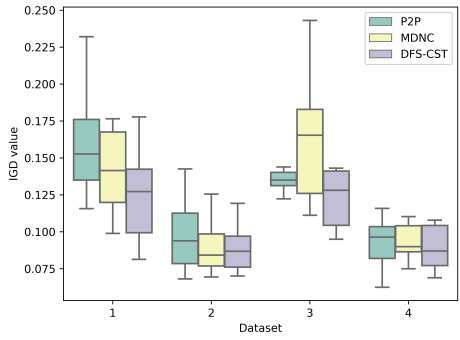}}
	\subfigure[datasets 13-16]{\includegraphics[width=0.43\textwidth]{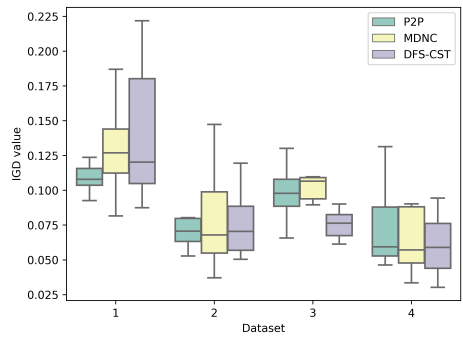}}
	\caption{Box plots of IGD  of each algorithm on synthetic workflow.}
	\label{fig:igd}
\end{figure*} 

\subsubsection{Results and Analysis based on IGD and HV}
Based on the same multi-objective optimization framework, we compare three clustering algorithms, i.e., P2P, MDNC, and DFS-CST.
First, the IGD values obtained from the three algorithms in 10 runs are recorded for each workflow set. The box plots in 16 datasets are shown in Figure~\ref{fig:igd}. In this figure, the x-axis denotes the numbering of datasets and the y-axis is the IGD values of different algorithms.
From these figures, it can be seen that in most cases, MDNC holds the largest value, and the IGD values obtained by P2P are greater than the proposed methods, which indicates that the proposed method is the best compared to the compared methods.   

\begin{figure*}
	\centering
	\subfigure[datasets 1-4]{\includegraphics[width=0.43\textwidth]{FIG_list/hv_for_dataset_1_three_method.pdf}}
	\subfigure[datasets 5-8]{\includegraphics[width=0.43\textwidth]{FIG_list/hv_for_dataset_2_three_method.pdf}}
	\subfigure[datasets 9-12]{\includegraphics[width=0.43\textwidth]{FIG_list/hv_for_dataset_3_three_method.pdf}}
	\subfigure[datasets 13-16]{\includegraphics[width=0.43\textwidth]{FIG_list/hv_for_dataset_4_three_method.pdf}}	
	\caption{Box plots of HV  of each algorithm on synthetic workflow.}
	\label{fig:hv}
\end{figure*} 

In addition to IGD, we also compare the HV values obtained by different methods. The reference point used in computing each HV value is [1.1, 1.1, 1.1], and each optimization objective is converted to a value between 0 and 1 before calculating HV values. We take the HV values in 10 runs as comparison values. The results for each dataset are shown in Figure~\ref{fig:hv}. In each figure, the x-axis indicates the numbering of datasets and the y-axis is the HV values obtained by different algorithms.
From these figures, it can be seen that the DFS-CST-based clustering methods still perform better in most cases. 

\subsubsection{Results and Analysis based on RDI}
Other than comparing the boxplots of IGD and HV, the Relative Deviation Index (RDI) of the mean of IGD and HV is also calculated.
RDI was defined by $RDI=(R_{current}-R_{best})/R_{best}$, where $R_{current}$ is the value obtained by the current algorithm and $R_{best}$ is the best value within all solutions obtained by all comparison algorithms~\citep{li2015elastic}.
The value can be IGD mean (RDI for IGD) or HV mean (RDI for HV).
It is noticed that the smaller the better for IGD and opposite for HV.
Hence, for IGD, the best value is the smallest solution among all algorithms while the best value refers to largest solution for HV.
The RDI results obtained in 16 datasets are shown in Table~\ref{RDI-metric} and the best value in each dataset is in bold.
From this table, it can be seen that RDIs of IGD mean which is obtained by DFS-CST hold the smallest values in most cases while RDIs of HV mean possess the largest values in most cases.
It further indicates that DFS-CST achieves the best results with the smallest RDI for IGD and the largest RDI for HV in most datasets.

\begin{table*}
	\footnotesize
	\centering
	\caption{RDI value for IGD mean and HV mean}
	\begin{tabular}{c|c|c|c|c|c|c}
		\hline
		\hline
		\multirow{2}{*}{Datasets} & \multicolumn{3}{c|}{RDI for IGD mean value} & \multicolumn{3}{c}{RDI for HV mean value}\\
		\cline{2-7}
		& P2P & MDNC & DFS-CST& P2P & MDNC & DFS-CST\\
		\hline
1	&	0.138	&	0.645	&	\textbf{0.000}	&	-0.008	&	-0.115	&	\textbf{0.000}	\\
2	&	0.105	&	\textbf{0.000}	&	0.215	&	-0.043	&	\textbf{0.000}	&	-0.087	\\
3	&	0.101	&	0.761	&	\textbf{0.000}	&	-0.005	&	-0.136	&	\textbf{0.000}	\\
4	&	0.148	&	0.387	&	\textbf{0.000}	&	-0.061	&	-0.069	&	\textbf{0.000}	\\
5	&	0.120	&	0.410	&	\textbf{0.000}	&	-0.055	&	-0.117	&	\textbf{0.000}	\\
6	&	\textbf{0.000}	&	0.066	&	0.007	&	-0.034	&	\textbf{0.000}	&	-0.001	\\
7	&	0.089	&	0.168	&	\textbf{0.000}	&	-0.099	&	-0.112	&	\textbf{0.000}	\\
8	&	0.145	&	0.145	&	\textbf{0.000}	&	-0.020	&	-0.064	&	\textbf{0.000}	\\
9	&	0.119	&	0.160	&	\textbf{0.000}	&	-0.033	&	-0.036	&	\textbf{0.000}	\\
10	&	0.080	&	0.115	&	\textbf{0.000}	&	-0.059	&	-0.023	&	\textbf{0.000}	\\
11	&	\textbf{0.000}	&	0.201	&	0.150	&	-0.156	&	\textbf{0.000}	&	-0.154	\\
12	&	0.009	&	\textbf{0.000}	&	0.035	&	-0.035	&	\textbf{0.000}	&	-0.032	\\
13	&	\textbf{0.000}	&	0.205	&	0.262	&	-0.016	&	\textbf{0.000}	&	-0.190	\\
14	&	0.017	&	0.015	&	\textbf{0.000}	&	-0.074	&	-0.027	&	\textbf{0.000}	\\
15	&	0.262	&	0.420	&	\textbf{0.000}	&	-0.086	&	-0.108	&	\textbf{0.000}	\\
16	&	0.204	&	0.175	&	\textbf{0.000}	&	-0.133	&	-0.040	&	\textbf{0.000}	\\
		\hline
	\end{tabular}
	\label{RDI-metric}
\end{table*}

Hence, based on different comparison perspectives mentioned above, it can be concluded that the DFS-CST-based approach outperforms the P2P and MDNC methods in most cases. The possible reason is that P2P can only cluster very few tasks together and cannot significantly reduce the communication time, which further increases the completion time of workflows, whereas MDNC clusters more tasks and allocates them to the same resource. Thus, the tasks lose the opportunity to be allocated to some more efficient resources. However, DFS-CST can cluster the suitable number of tasks, more than P2P and less than MDNC. This enables DFS-CST cluster more tasks than P2P, which can reduce communication time, and fewer tasks than MDNC, which allow tasks to be more flexible in selecting resources. 

\section{Conclusion} \label{sec:conclusion}

This paper focuses mainly scheduling multiple workflows on the cloud considering multiple objectives.
This work is the first attempt to study the multi-objective optimization model from the perspectives of the whole makespan and cost as well as the individual fairness.
A new definition of fairness among workflows  is proposed, which is based on individual makespan and individual cost.
For the multi-objective scheduling problem of multiple workflows, a DFS-CST-based clustering method is presented and compared with other existing global-based clustering methods.
The superiority of the proposed method in obtaining the better solutions for the considered objectives are discussed. 
The obtained results support the execution of computing tasks in distributed environments from an algorithmic perspective.

There are several aspects that need further investigation.
For instance, resources can be considered dynamic in the processing of scheduling.
The scheduling results should also be validated in real-world.  
\section*{Funding}

This work was supported by the A*STAR Cyber-Physical Production System (CPPS) – Towards Contextual and Intelligent Response Research Program, under the RIE2020 IAF-PP Grant A19C1a0018, and Model Factory@SIMTech

\bibliographystyle{tfcad}
\bibliography{myRef}

\end{document}